\documentclass[10pt, a4paper]{article}
\usepackage{lrec2022} % this is the new LREC2022 Style
\usepackage{multibib}
\newcites{languageresource}{Language Resources}
\usepackage{graphicx}
\usepackage{tabularx}
\usepackage{soul}
\usepackage{amssymb}
\usepackage{titlesec}
\titleformat{\section}{\normalfont\large\bfseries\center}{\thesection.}{1em}{}
\titleformat{\subsection}{\normalfont\SmallTitleFont\bfseries\raggedright}{\thesubsection.}{1em}{}
\titleformat{\subsubsection}{\normalfont\normalsize\bfseries\raggedright}{\thesubsubsection.}{1em}{}
\renewcommand\thesection{\arabic{section}}
\renewcommand\thesubsection{\thesection.\arabic{subsection}}
\renewcommand\thesubsubsection{\thesubsection.\arabic{subsubsection}}
\usepackage{epstopdf}
\usepackage[utf8]{inputenc}

\usepackage{hyperref}
\usepackage{xstring}

\usepackage{color}

\usepackage{amsmath}
\DeclareMathOperator*{\argmin}{arg\,min}
\usepackage{graphicx}

\title{Probing Pre-trained Auto-regressive Language Models for Named Entity Typing and Recognition}
\name{Elena V. Epure, Romain Hennequin} 

\address{Deezer Research \\ 22-26 rue de Calais, 75009 Paris, France \\
    research@deezer.com\\}

\abstract{
Multiple works have proposed to probe language models (LMs) for generalization in named entity (NE) typing (NET) and recognition (NER). 
However, little has been done in this direction for auto-regressive models despite their popularity and potential to express a wide variety of NLP tasks in the same unified format. We propose a new methodology to probe auto-regressive LMs for NET and NER generalization, which draws inspiration from human linguistic behavior, by resorting to meta-learning. We study NEs of various types individually by designing a zero-shot transfer strategy for NET. Then, we probe the model for NER by providing a few examples at inference. We introduce a novel procedure to assess the model's memorization of NEs and report the memorization's impact on the results. Our findings show that: 1) GPT2, a common pre-trained auto-regressive LM, without any fine-tuning for NET or NER, performs the tasks fairly well; 2) name irregularity when common for a NE type could be an effective exploitable cue; 3) the model seems to rely more on NE than contextual cues in few-shot NER; 4) NEs with words absent during LM pre-training are very challenging for both NET and NER. \\ 
\newline \Keywords{auto-regressive language model, probing, zero-shot NET, few-shot NER, memorization testing}}

\begin{document}

\maketitleabstract

\section{Introduction}
\label{sec:intro}
Before transformer LMs \cite{devlin-etal-2019-bert}, the state-of-the-art NER was based on training recurrent neural networks, such as bidirectional LSTM (BiLSTM) with a Conditional Random Field (CRF) layer, from scratch \cite{yadav-bethard-2018-survey}.
The widely adopted approach with transformers has been to fine-tune them for the desired task, thus specializing their general linguistic knowledge, acquired during pre-training.
While LMs fine-tuned for NER have achieved impressive results on standard benchmarks \cite{akbik-etal-2019-pooled}, multiple works have emphasized their limitations with regard to their generalization capacity to new textual genres (e.g. clean versus noisy text), NE type sets (e.g. NE types belonging to new domains such as music or e-commerce) and new NEs, unseen during training \cite{lin-etal-2020-rigorous}.

To gain more insights into the NER generalization ability of LMs,  multiple studies have been conducted. 
Probing has been designed for BiLSTM-CRF LMs \cite{Augenstein2017Generalisation,Taille2020contextualized,fu2020rethinking} or masked LMs such as BERT \cite{petroni-etal-2019-language,jiang-etal-2020-know}.
Yet, little has been done for auto-regressive models such as GPT2 despite their popularity and potential to express a wide variety of NLP tasks in the same unified format \cite{Raffel2020Exploring}.

Additionally, although the past probing studies have broadened the knowledge about how LMs generalize in the NER context, multiple improvements could be brought to existing methodologies.
First, the impact of pre-training LMs on the results has never been assessed.
%However, when investigating a LM by fine-tuning it for NER, the knowledge the LM already has via pre-training may also account for a part of the performance. 
Second, the proposed setups test generalization by relying on large annotated datasets which are manipulated in different ways to create test and train splits.
However, when assessing generalization in relation to \textit{human linguistic behavior} \cite{levesque2014our}, which we claim as more realistic, these datasets are insufficient and different testing conditions should exist. 
Humans can easily recognize NEs based on prior domain and common sense linguistic knowledge, or by leveraging contextual cues in text \cite{lin-etal-2020-triggerner}.
Humans can perform new linguistic tasks quite well even when exposed to a few examples or very simple instructions \cite{Brown2020Language}.
When it comes to technology creation, human linguistic behavior could lead to infinite examples, many of them new to everyone, including to the systems' designers  \cite{emnlp-2020-2020}.

Hence, datasets used in past studies cannot capture this variability for a realistic testing unless continuously updated.
%We propose a probing methodology for NER generalization targeting auto-regressive LMs by considering the human linguistic behavior and humanlike conditions.
However, like humans, LMs have gained diverse domain and linguistic knowledge, and developed general pattern recognition abilities from experience, during pre-training \cite{Brown2020Language}.
Given these, the research question we investigate is: 

Can the knowledge gained during pre-training be leveraged by auto-regressive LMs at inference to adapt to diverse NE-related tasks, when queried with a few examples at most and simple natural language instructions?

\paragraph{Contributions.} 
Inspired by testing conditions related to human linguistic behavior, we design a probing methodology centered on meta- or "in-context" learning.
It entails the task specification  via the text input used to prompt the model, without performing any gradient updates \cite{Brown2020Language}.
First we study NEs of various types individually by defining a zero-shot transfer strategy for NET.
We design \textit{a novel method to assess NE memorization} by the model and report the memorization's impact on the results.
Our memorization method could be used to sample (un)popular NEs \cite{shwartz-etal-2020-grounded}, but also, beyond the NE context, with other types of n-grams.
Second, we model NER as a machine reading comprehension (MRC) task and probe the model by providing a few examples at inference (e.g., the model should extract spans of text from input, as answers to simple queries).
We also test NER with (un)memorized NEs and gain insights on the role of context, i.e. text around NEs.
We use four datasets: CoNLL-2003 \cite{tjong-kim-sang-de-meulder-2003-introduction}, WNUT2017 \cite{liu2014a}, MIT Movie \cite{liu2014a} and extensive lists of NEs from DBpedia \cite{Auer2007Dbpedia}.
These datasets contain clean and noisy text, and regular NEs such as people names and irregular NEs such as creative work titles.

Our study\footnote{Code is available at \href{https://github.com/deezer/net-ner-probing}{https://github.com/deezer/net-ner-probing}.} joins other efforts that looked into NET and NER generalization but that, compared to us, achieved this by manipulating datasets during fine-tuning / testing or targeted other types of LMs. 
To our knowledge, we are the first to extensively probe \textit{pre-trained auto-regressive LMs as they are} for these tasks and ensure testing conditions related to human linguistic behavior.

\paragraph{Findings.} 
Pre-trained GPT2, a common auto-regressive LM, appears to perform the tasks fairly well without any fine-tuning for NET or NER, especially on regular NEs or memorized during pre-training.
These models, as they are, already know quite a lot about NEs and encode NER patterns.
Our finding is particularly important given that past works study NET and NER generalization of existing LMs without explicitly considering the impact of model pre-training.
Then, compared to other studies that claim named entity irregularity to be problematic \cite{Augenstein2017Generalisation}, we show that when frequently present for a certain NE type it can become, in fact, an effective exploitable cue. 
We also show that the model seems to rely more on NE cues than on context cues in few-shot NER, and that the model’s exposure to the NE's words weighs much more than the exposure to the exact NE in zero-shot NET.

\section{Background and Related Work}
\label{sec:related}
\subsection{NE Generalization in Current Models}
The common way to perform NER nowadays relies on training or fine-tuning a deep neural network using a relatively large annotated dataset and often aims at extracting a few regular NE types such as person, location and organisation \cite{yadav-bethard-2018-survey,akbik-etal-2019-pooled,lison-etal-2020-named}.
Although recent LMs have yielded impressive results, NER generalization to all types of textual genres is still an issue, in particular, in informal text, frequently found on social media or in chat-bot interactions.
This type of text can often lack proper formatting, e.g. word capitalization, and contain unusual grammatical structures or jargon
\cite{aguilar-etal-2018-modeling,guerini-etal-2018-toward}.

Another challenge is NER generalization to diverse and growing NE type sets, belonging to new domains such as movies, music or e-commerce \cite{ma-etal-2016-label,guerini-etal-2018-toward,lin-etal-2020-rigorous}. 
These types are often more heterogeneous (e.g. \textit{groups} in WNUT includes sport teams and music bands \cite{aguilar-etal-2018-modeling}); lack name regularity (e.g. \textit{creative work} titles are not necessarily noun phrases \cite{lin-etal-2020-rigorous}); can be composed of common words or of words which are typically from other languages (e.g. the film "Demolition Man" \cite{derczynski-etal-2017-results}).
Then, NER generalization to new NEs, unseen during training is another challenge. 
This is common in the real-world where a system learns from a limited number of examples per type while NE mentions are expected to shift in time \cite{Augenstein2017Generalisation}.

These challenges have been addressed by relying on new training datasets with each new case. 
However, collecting thousands of human annotations for new genres, NE types or NE mentions is expensive and time-consuming \cite{Augenstein2017Generalisation,lin-etal-2020-triggerner}.
Other recent works rely on existing NE resources, such as gazetteers and dictionaries, to either perform NER in a distant or weak supervision setup \cite{lison-etal-2020-named,shang-etal-2018-learning}, or to train NET classifiers adaptable to unseen NEs \cite{guerini-etal-2018-toward}.
Constraining the model to rely more on context than on NEs have also appeared promising to achieve generalization \cite{mengge-etal-2020-coarse,lin-etal-2020-triggerner}. 

Other efforts towards NET and NER generalization share the same rationale as us---the challenges in collecting annotations with each new case and the human linguistic behavior, and design zero- or few-shot learners to perform the tasks \cite{zhou-etal-2018-zero,zhang-etal-2020-mzet,yang-katiyar-2020-simple,ding-etal-2021-nerd,aly-etal-2021-leveraging}.
There are several major differences with our study.
First, our goal is not to propose a new NET or NER model, but to probe pre-trained LMs as meta-learners without modifying their weights or leveraging external knowledge.
Second, assuming that our probing methodology is exploited as a basic NET and NER model, its input is much more constricted (lists of NEs and NE types in NET, a few NEs in context for NER) compared to the other works which make use of a wide range of resources such as:
knowledge bases, definitions of entity types in a taxonomic and/or natural language forms, NEs in context even for NET, or datasets of seen NE types with many examples.

\subsection{Probing Studies for NER Generalization}
\newcite{lin-etal-2020-rigorous} propose an extensive use of randomization tests to study the extent to which a fine-tuned LM relies on: name regularity---regular (e.g. persons) versus irregular names (e.g. creative works); on mention coverage---the ratio of overlapping NEs in train and test data; and on context diversity---unique sentences for each NE type.
\newcite{fu2020rethinking} investigate the popular NER architecture, LSTM-CRF, from various views including NE and contextual coverage.
Also, they study the impact of the relations among NE types on model learning.
A BiLSTM-CRF is also studied in \cite{Taille2020contextualized}, but the focus is on bench-marking different contextualized or static embeddings for generalization to new NE mentions and domains.

Compared to these, we focus on pre-trained auto-regressive LMs as-is and not on fine-tuning / training them and, implicitly, on the impact of train / test datasets.
To our knowledge, this is the first detailed study designed for pre-trained LMs as NET and NER meta-learners without modifying them or using external resources.
We study many generalization angles: seen versus unseen NEs--referred also as \textit{memorized versus unmemorized NEs} in the paper, regular and irregular NEs, diverse genres including noisy text, and reliance on context versus NE cues.

\section{Proposed Probing Methodology}
\label{sec:ner}
%The proposed methodology to probe LMs for NET and NER is inspired by human l who can perform these tasks when given simple instructions and at most few examples, and can generalize to unknown NEs \cite{kobeleva2012second}.
We divide our study in NET in a zero-shot transfer  followed by NER in a few-shot settings.
This allows us to acquire knowledge first about names and then about NEs in context.

\subsection{NET in a Zero-shot Setup}
\label{sec:zero}

\begin{figure*}[htb]
\begin{small}
    \centering
    \includegraphics[width=.32\linewidth]{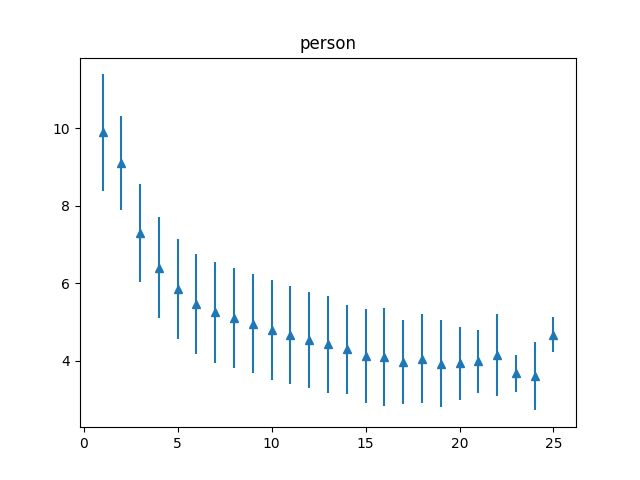}
    \includegraphics[width=.32\linewidth]{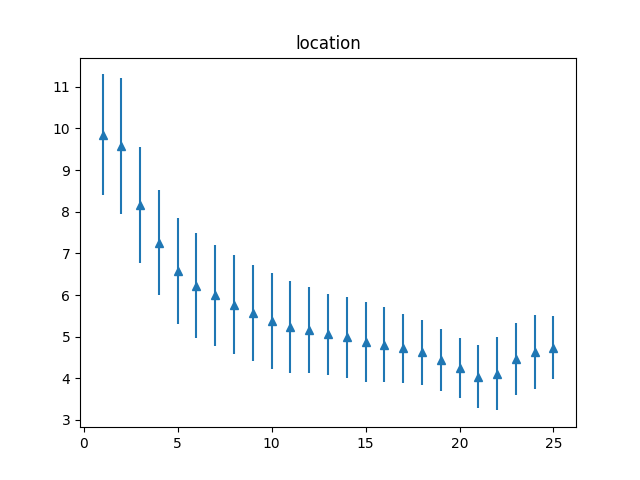}
    \includegraphics[width=.32\linewidth]{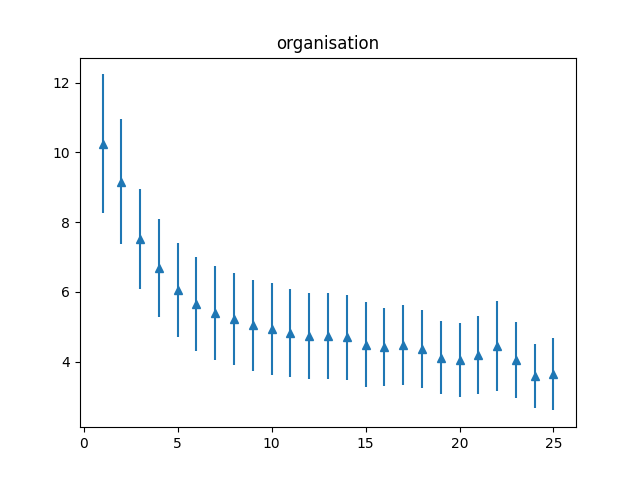}
\caption{Mean and standard deviation of log perplexity computed with GPT2 for large lists of persons (left), locations (middle) and organisations (right) from DBpedia. 
    Values are grouped by the number of tokens per NE.
}
    \label{fig:ppltoks}
\end{small}
\end{figure*}

Auto-regressive LMs estimate the empirical distribution from the training data, where each training example $x$ is a sequence of tokens $x=(s_1, s_2, ..., s_n)$. 
Given the sequential nature of the language, it is common to factorize the distribution $p(x)$ with the Bayes' rule and express it as a product of conditional probabilities of each sequence's token $s_i$ given the previous tokens:
\begin{equation}
    p(x) = \prod_{i=1}^{n}{p(s_i|s_1,...,s_{i-1})}
\end{equation}
On a new task, the model infers $p(output|input)$ or more completely written $p(output|input, task)$.
\newcite{Brown2020Language} merge the $input$ and $task$ in a single natural language query and express $output$ as the predicted next sequence of tokens.
For instance, we could write a query for NET as "Sentence: is Italy a person, location or organisation? Answer:".
The predicted NE type is then the token among "person", "location", or "organisation" which the model estimates as most likely to follow.

Alternatively, we could frame NET as the most likely statement among multiple competing ones such as "Anne is a person", "Anne is a location".
In this case, the sequence with the lowest perplexity is the one that the model is less surprised to see, hence describing the most likely NE type.
The perplexity of a sequence $x$, using a model $\theta$, is:
\begin{equation}
    \text{PPL}_\theta(x)=\exp\{-\frac{1}{n}\sum_{i=1}^{n}{\log p_\theta}(s_i|s_{<i})\}
\end{equation}
We adopt this latter approach as it provides a simple task framing in zero-shot settings and perplexity can be efficiently computed by relying on a single model call.
Thus, given a NE mention $e$ and a NE type set $T$, the most likely type $t_e \in T$ for $e$ is:
\begin{equation}
t_e = \argmin_{t \in T}{\text{PPL}_\theta(\text{query}(e,t))}
\end{equation}
where $\text{query}(e,t)$ is the template "$e$ is a $t$" (e.g. "Cinderella is a city" or "Cinderella is a character").

\paragraph{Assessing generalization.} The model's generalization to NEs unseen or rare during training is essential for a human-centered setup.
Thus, assessing how the model performs on (un)memorized NEs could provide a more realistic understanding of its performance.
Previous works that led such investigation, trained models from scratch \cite{lin-etal-2020-rigorous,Taille2020contextualized}, so could keep track of (un)seen NEs during training.
As we focus on pre-trained LMs and have no access to their training data, we devise a method to assess if the model has memorized a NE or not.

\newcite{Carlini2019Secret} propose a test for unintended memorization of rare sequences based on perplexity.
Given all possible sequences for a matter at hand (or a very large sample, $S$) prefixed by the same query (e.g. prefix "the random number is " and $S=\{281265011, 281265017 ...\}$), rank them by PPL$_\theta$ and use ranks to compute exposure:

\begin{equation}
    \text{exposure}_\theta(x) = \log_2{|\mathcal{S}|}-\log_2{\text{rank}_\theta(x)}
    \label{eq:exposure}
\end{equation}
\noindent For $x\in\mathcal{S}$, the exposure metric is negatively correlated with the rank, i.e. the lower the rank the higher the exposure, thus likely memorization.

This test is a helpful point of departure, but less applicable to our task as-is.
Without a very large set of NEs for each type, the estimates could be inaccurate, especially when only few sequences have lower perplexity than a target one \cite{Carlini2019Secret}.
Also, we noticed experimentally (see Figure \ref{fig:ppltoks}) that the mean perplexity tended to decrease with the number of tokens per NE\footnote{The trend was similar per number of words or characters.}, a phenomenon most likely related to the open-vocabulary language modeling over sub-word units\footnote{Among the NEs with large number of tokens and lower perplexity, we often noticed NEs from other languages than English and with more words (e.g. \emph{L'Hospitalet-près-l'Andorre}).
Given the previous ones, the probability of many of these sub-tokens is quite high (e.g. $p(\textnormal{s}|\textnormal{L'Hospitalet-prè})=0.993$), which results in low perplexity. 
This could be an effect of the model memorizing some rare or foreign words, but not necessarily the NE.}.
Thus, with the method of \newcite{Carlini2019Secret}, NEs would have a higher chance to be flagged as memorized when they are tokenized in more tokens. 

As originally stated, our goal is to evaluate the model's behavior with (un)memorized NEs.
Thus, we want to be able to assign NEs to two groups when we are confident of their (non-)memorization, while ignoring NEs in the gray area.
We changed the previously shown test to rely directly on probabilities of NE's tokens, obtained when calling the model with NEs as input, prefixed by a fixed string.
The test we propose is further summarized:

If NE \textit{words} are known (e.g. "Great" and "Britain" are in the model's vocabulary) and their sequential \textit{transitions} are unsurprising (e.g. $p(\textnormal{Britain}|\textnormal{Great})$ is large), then the model has likely seen the NE during training.
Formally, we define two exposure metrics for these two aspects as follows:

\begin{equation}
\begin{split}
\text{exposure}_{\theta}^{word}(x) = \prod_{(i,j) \in W_x}{\text{test}_{\theta}^{word}(x,i,j)} \\
\text{test}_{\theta}^{word}(x,i,j) =
\left\{
\begin{array}{ll}
    1 & \mbox{if } i = j \\
    {p_\theta}(s_{j}|s_{<{j}}) & \mbox{if }  i < j
\end{array}
\right.
\end{split}
\label{eq:exposure2}
\end{equation}

\begin{equation}
\text{exposure}_{\theta}^{trans}(x) =
\min_{i \in T_x}{p_\theta}(s_{i}|s_{<{i}})
\label{eq:exposure3}
\end{equation}
where $W_x$ consists of tuples marking the start and end indices of each word in $x$ (a word can have multiple tokens) and $T_x$ has indices marking the transitions (the index of each new word). 
In Equation \ref{eq:exposure2}, we identify two cases when a word can be considered known by the LM.
It is directly mapped on a token in the LM's vocabulary $V$ or, when it is split in multiple tokens, the last token becomes an indicator of its memorization.
In Equation \ref{eq:exposure3}, to test whether the sequential association of words is unsurprising to the model, we take the minimum probability of the tokens marking the start of each new word.

For the final decision, NEs with exposure values higher or lower than some established thresholds could be assigned to the memorized / unmemorized NE groups.
These thresholds could be defined considering the NE set and the model's vocabulary size (more details in Section \ref{sec:experiments}). 
An advantage of our method over \cite{Carlini2019Secret} is that we do not need access to a very large set for each NE type, the token probabilities being sufficient to establish the NE exposure / degree of memorization.

\subsection{NER in a Few-shot Setup}
\label{sec:few}
We frame NER as a MRC task \cite{mengge-etal-2020-coarse,li-etal-2020-unified}, but, instead of fine-tuning / training a pre-trained LM for MRC, we exploit it in a few-shot setup.
As detailed in Section \ref{sec:intro}, the few-shot setup has been considered closer to human linguistic behavior and has shown competitive results in other NLP tasks such as question answering, translation, and classification \cite{Brown2020Language}.
\newcite{Zhao2021Calibrate} have also tested information extraction for slot-filling with some slots targeting NEs (e.g. the director of a movie).
Yet, they assume that each sentence contains that type of slot, without assessing the case when no NEs exist in the sentence. 

Similar to past works, we use a query to formulate the task and insert examples, which are used only at inference, without triggering updates of the pre-trained model weights.
We show in Figure \ref{fig:prefix} a query generated from WNUT2017 dataset for the entity type \textit{product}.
The query has two parts: a prefix and a test sentence.
The \textit{prefix} is appended to each sentence and introduces the examples (0-7).
We provide $4$ examples, two with NEs and two without.
We use the token "none" to mark the absence of a NE of the target type.
The second part with the last two lines  (8-9) introduces the \textit{test sentence}, for which NER is performed.

\begin{figure}[htb]
\begin{small}
    \centering
    \begin{tabular}{cll}
        0: & Sentence: & I don't like to be stuck at home \\
        1: & product: & none \\
        2: & Sentence: & Where is Gelato Gilberto? \\
        3: & product: & none \\
        4: & Sentence: & Well, I was gonna buy a Zune HD \\
        5: & product: & Zune HD \\
        6: & Sentence: & BEAUTY TIPS: SK-II UV Cream \\
        7: & product: & SK-II UV Cream \\
        8: & Sentence: & CVS sells their own epipen \\
        9: & product: & \\ \hline
        & \textit{True Answer}: & epipen
    \end{tabular}
    \caption{NER query and the expected generated answer. Lines 0-7 are examples (two negatives, two positives) and lines 8-9 are the test sentence. }
    \label{fig:prefix}
\end{small}
\end{figure}

Previous works have shown that the query choice has a significant impact on the task's accuracy \cite{li-etal-2020-unified}.
In the few-shot learning case, the set of examples and their order can lead to different results too \cite{Zhao2021Calibrate}.
For instance, the model might tend to predict the majority token or the one nearest to the end of the query.
To overcome this, \newcite{Zhao2021Calibrate} propose a procedure to calibrate the LM's output probabilities by taking into account the LM's bias towards certain outputs.
Specifically, in the generation task, an affine tranformation and softmax is applied to $\hat{\textnormal{\textbf{p}}}$, the set of probabilities of the first token: $\hat{\textnormal{\textbf{p}}}_{cal} = \textnormal{softmax}(\textnormal{\textbf{W}}\hat{\textnormal{\textbf{p}}})$.
$\textnormal{\textbf{W}}$ is estimated from $\hat{\textnormal{\textbf{p}}}_{cf}$, the probabilities obtained when feeding in the model a "content-free" input such as "N/A", as $\textnormal{\textbf{W}}=\textnormal{diag}(\hat{\textnormal{\textbf{p}}}_{cf})^{-1}$.
We use the same calibration procedure and run each experiment multiple times, with varied examples as demonstration.
As for the query format, we stick to the one shown in Figure \ref{fig:prefix} and leave for future work the exploration of other formats.

\begin{table*}
    \centering
    \begin{tabular}{c|c|c|cc}
    \textbf{Dataset} & \textbf{Type} & \textbf{NE types} & \textbf{NET} & \textbf{NER} \\ \hline
     CoNLL-2003  & clean & person, location, organisation & $\checkmark$ & $\checkmark$ \\
     MIT Movie & noisy & person, creative work &  $\checkmark$ & $\checkmark$ \\
     WNUT2017 & noisy & person, location, corporation, group, product, creative work &  $\checkmark$& $\checkmark$ \\
     DBpedia & clean & person, location, organisation, creative work & $\checkmark$ & \\ \hline
    \end{tabular}
    \caption{Overview of the datasets used in each task. For NET, we consider NE mentions from all dataset (train, test, and dev if available). NER is evaluated on test sets while the train sets are used only to sample examples for the query.}
    \label{tab:datasets}
\end{table*}

\paragraph{Assessing generalization.}
Assessing if the LM can generalize to unmemorized NEs is a more humanlike setup for evaluation.
We could select two sets of sentences with our memorization test, and report performance on each separately.
However, splitting smaller datasets as WNUT2017 would not allow for reliable conclusions, nor to analyse the role of the context (i.e. the other parts of the sentence) because sentences would be different in the dataset splits.
As \newcite{lin-etal-2020-rigorous} highlighted, we should aim at NER models that rely more on context for generalization, rather than memorizing NEs, in particular for irregular NE types such as creative work titles.
Thus, it is relevant to allow for the study of context.

For these reasons, the experiment design we propose is to fix the examples in the query prefix (0-7 in Figure \ref{fig:prefix}), and compute performance on three variations of the test sentence (8-9 in Figure \ref{fig:prefix}): \textit{test as-is} (the original sentence), \textit{test seen} and \textit{test unseen}.
To obtain \textit{test seen}, we randomly sample NEs to replace the existing ones in the original sentence from a list of memorized NEs identified with our memorization test.
For \textit{test unseen}, we replace NEs by choosing among random lowercase strings, which do not exist in the English language, thus are not memorized.
The hypotheses are: 1) if the model relies more on context for NER then the performance on \textit{test as-is} and \textit{test unseen} should be similar; 2) if the model relies more on NE cues then the performance on \textit{test seen} should be much larger than on \textit{test as-is}.

\section{Experiments}
\label{sec:experiments} 
We apply the proposed methodology to a medium-sized GPT2.
A larger model like GPT3 yielded better results as a meta-learner in past experiments \cite{Zhao2021Calibrate}.
However, we have decided to use the proposed probing methodology with a model that was easily accessible and had lower memory requirements, leaving the extension to other auto-regressive models as future work.

\paragraph{Datasets.} 
CoNLL-2003 \cite{tjong-kim-sang-de-meulder-2003-introduction} and WNUT2017 \cite{derczynski-etal-2017-results}, commonly found in NER benchmarks, are kept as they are.
The MIT Movie dataset \cite{liu2014a}, originally created for slot filling, is modified by ignoring some slot types (e.g. \textit{genre}, \textit{rating}) and merging others (e.g. \textit{director} and \textit{actor} in \textit{person}, and \textit{song} and \textit{movie title} in \textit{title}) in order to keep consistent NE types across all datasets.
MIT movie dataset contains only lowercase text, sometimes with typos, thus falling under the noisy text genre as WNUT2017.
For NET, we consider the NE mentions from each dataset in its entirety (train, test, and dev if available).
We also collected large lists of different NE types from DBpedia \cite{Auer2007Dbpedia}.
These are particularly interesting because Wikipedia has not been included in the GPT2 training corpora \cite{radford2019language}.
The NER experiments are run only on test sets while the train sets are used for sampling examples for the query.
A summary of the datasets and the tasks in which they are used is presented in Table \ref{tab:datasets}.
\begin{table*}
\begin{small}
    \centering
    \begin{tabular}{c|c|ccc|cc|cc}
      \textbf{Dataset} & \textbf{NE Type} & \multicolumn{3}{c|}{\textbf{All}} & \multicolumn{2}{c|}{\textbf{Memorized}} & \multicolumn{2}{c}{\textbf{Unmemorized}} \\ 
      & & \textbf{F1} & \textbf{F1 (ZOE)} & \textbf{Count} & \textbf{F1} & \textbf{Count} & \textbf{F1} & \textbf{Count} \\\hline
      CoNLL-2003 & person & 0.90 & 0.90 & 3613 & 0.93 & 695 & 0.86 & 619 \\
      & location & 0.66 & 0.80 & 1331 & 0.74 & 546 & 0.37 & 80 \\
     & organisation & 0.70 & 0.74 & 2401 & 0.74 & 770 & 0.63 & 289 \\
       & \textit{macro-average} & 0.75 & 0.81  &   7345 & 0.81 & 2011 & 0.62 & 988 \\ \hline
      MIT Movie & person & 0.80 & - &  2866   & 0.82 & 605 & 0.81 & 369 \\
        & creative work & 0.60 & - &  2122   & 0.64 & 402 & 0.58 & 256 \\
        & \textit{macro-average} & 0.70 & - &  4988  & 0.73 & 1007 & 0.69 & 625 \\ \hline
    \end{tabular}
    \caption{Zero-shot NET F1-scores. Results obtained with ZOE on CoNLL-2003 are also shown.} 
    \label{tab:entitytyping}
\end{small}
\end{table*}

\begin{table}
\begin{small}
    \centering
    \begin{tabular}{c|c|ccc}
   \textbf{Metric} & \textbf{NE Type} & \textbf{M} & \textbf{UM}
    & \textbf{Count} \\ \hline
    $\text{exposure}_{\theta}^{word}$ & person  & 0.88 & 0.64 & 10000 \\
    & location  & 0.81 & 0.63 & 10000 \\
    & organisation  & 0.76 & 0.67 & 10000 \\
    & creative work & 0.69 & 0.36 & 10000 \\ 
    & \textit{macro-average} & 0.78  & 0.58 & 40000 \\ \hline  
    $\text{exposure}_{\theta}^{tran}$ & person  & 0.83 & 0.78 & 10000\\
    & location  & 0.83 & 0.75 & 10000\\
    & organisation  & 0.78 & 0.71 & 7014 \\
    & creative work & 0.63 & 0.55 & 6012 \\ 
    & \textit{macro-average} & 0.77  & 0.70 & 33026 \\ \hline
    \end{tabular}
    \caption{Zero-shot NET results on DBpedia NEs. M stands for Memorized and UM for Unmemorized. The \textbf{Metric} column reports the exposure metric used to select memorized and unmemorized NE lists.}
    \label{tab:dbp_entitytyping}
\end{small}
\end{table}

\paragraph{Probing NET.}
For NET, we create prompts starting from NE types and choose as predicted value the type which leads to the lowest perplexity as presented in Section \ref{sec:zero}.
In practice, we use multiple keywords for each NE type starting from their definition.
We also include \textit{character} for \textit{person}; \textit{company}, \textit{group}, \textit{institution}, \textit{club}, and \textit{corporation} for \textit{organization}; \textit{place}, \textit{city}, and \textit{country} for \textit{location}.
As the perplexity decreases with the number of tokens as shown in Figure \ref{fig:ppltoks}, we choose all keywords such that they are part of the model vocabulary.
Thus, \textit{creative work} is replaced by \textit{work}, \textit{title}, \textit{movie}, \textit{song}, and \textit{book}.
We do not include other keywords for \textit{product}, \textit{corporation} and \textit{group} in WNUT2017.

For the exposure computation, we prefix NEs with the default unknown token when retrieving probabilities.
The thresholds for word and transition exposures are established per dataset.
For the lower limit, we consider the size of the GPT2 vocabulary ($\approx50K$); thus, assuming a uniform word distribution\footnote{This assumption is strong, but used only to establish an order of magnitude for the unmemorized $\text{exposure}_{\theta}^{trans}$.}, each token would have a $2e\text{-}05$ probability to be generated next.
CoNLL-2003 has many one-word NEs with rare transitions.
For this reason, we focus only on $\text{exposure}_{\theta}^{word}$ to establish if a NE is memorized ($\geq.8$) or not ($\leq 1e\text{-}04$).
The rest of NEs are not classified.
In contrast, in MIT Movies, NEs are often composed of multiple words common in English-language, thus present in the model's vocabulary.
In this case, $\text{exposure}_{\theta}^{tran}$ is more informative for selecting memorized NEs ($\geq.001$) and unmemorized NEs ($\leq 1e\text{-}05$).

We sample the two groups from  the DBpedia lists using either $\text{exposure}_{\theta}^{word}$ or $\text{exposure}_{\theta}^{tran}$.
In this way, we investigate the impact of knowing words vs. recognizing word transitions on a much larger sample.
We ignore one-word NEs from MIT Movies and DBpedia because they are rare or often spurious.
Finally, we only run the NET experiment on the complete WNUT2017 dataset because the number of NEs for each entity type is too small to allow reliable memorized vs. unmemorized split.

\paragraph{Probing NER.} We opt for a maximum of training examples in the query that can be kept in memory, in our case $16$.
Out of these, $9$ contain NEs of the targeted type and $7$ are randomly chosen from the rest of the dataset.
We run each experiment three times with different random seeds to compute variance.
The test set is slightly modified too: for each NE type, we keep all positive sentences and sample negative sentences such that the ratio positive-negative is about 2:1. 
The maximum number of tokens asked when querying the model is set to $15$.
The calibration we apply follows the steps described in \cite{Zhao2021Calibrate}.

We design the NER meta-learner to extract one NE of the prompted type at a time, leaving the case of multiple NEs per text as future work.
Because a test sentence can mention multiple NEs of the same type, we consider a generated answer to be correct if it matches one of the existing NEs.
In computing scores, we rely mostly on exact NE matching with some exceptions.
The evaluation is insensitive to the letter case (e.g. 'none' and 'None' are considered equivalent).
Also, we noticed that the model tends to add spaces for NEs written together such as in social media mentions.
To cover these cases, we consider that the prediction is equal to the ground-truth, if their Levenshtein distance divided by the true NE length is lower than $0.2$.
When no NEs should be extracted but the model generates another string that does not have any words in common with the input, we consider it a correct prediction even if it's not explicitly "none"\footnote{The model can generate strings such as \textit{null} or "\textit{.}"}.

\paragraph{Validation.} As previously mentioned, out goal is not to propose a new NET or NER model, but to probe pre-trained LMs as meta-learners without modifying their weights or leveraging external knowledge.
However, as a sanity check to understand if the model used in this way actually works, we position the model's performance with respect to the performance of other baselines.

For NET on CoNLL-2003, we include a zero-shot baseline, ZOE \cite{zhou-etal-2018-zero}, which derives NE types by having as input the NE mention in a sentence and a taxonomy of NE types with the corresponding definition for each type. 
ZOE is designed for clean text and uses Wikipedia as the taxonomy of NE types.
For these reasons, we cannot apply it to the other noisy datasets or to DBpedia, which is extracted from Wikipedia.
Nonetheless, we also report the performance of a weighted random guess NET classifier for WNUT2017.

For NER, we report the results of the best supervised baseline of the shared WNUT2017 task, UH-RiTUAL \cite{aguilar-etal-2017-multi}.
UH-RiTUAL relies on a neural network to extract feature representations that are further fed in a CRF classifier.
The feature extractor model is trained in a multi-task fashion on two objectives, NE segmentation and NE classification, and leverages as input character embeddings, Part-of-Speech tag embeddings, word embeddings and gazetteers.

A recent comprehensive study on few-shot NER \cite{huang2020few} benchmarks multiple methods entailing training (prototype-based, self-training).
We also report their score range for each dataset (CoNLL-2003, MIT Movie and WNUT2017) and compare them to our results.

\section{Results and Discussion}
\label{sec:results}

\begin{table}
\begin{small}
    \centering
    \begin{tabular}{c|ccc}
    \textbf{NE Type} & \textbf{F1} & \textbf{F1 (random guess)} & \textbf{Count} \\\hline
    person & 0.79 & 0.40 & 1317 \\ 
       location & 0.63 & 0.19 & 616\\
       corporation & 0.16 & 0.07 & 231 \\
       group & 0.44 & 0.13 & 412 \\
       product & 0.46 & 0.11  & 353\\
       creative work & 0.46  &  0.11 & 361\\
       \textit{macro-average} & 0.49 & 0.17 & 3290 \\ \hline
    \end{tabular}
    \caption{Zero-shot NET F1-scores on WNUT2017. }
    \label{tab:wnut_entitytyping}
\end{small}
\end{table}

\begin{table*}
\begin{small}
    \centering
    \begin{tabular}{c|cc|cc|c}
      \textbf{NE type} & \textbf{Test as-is} & \textbf{UH-RiTUAL} & \textbf{Test seen} & \textbf{Test unseen}  &  \textbf{Count} \\ \hline
      person & 0.68$\pm$0.04 &  0.68 & 0.81$\pm$0.02 & 0.63$\pm$0.15 & 490 \\
      location & 0.67$\pm$0.01 &  0.71 & 0.81$\pm$0.04 & 0.61$\pm$0.07 & 187  \\
      corporation & 0.66$\pm$0.03 &  0.36 & 0.82$\pm$0.03 & 0.51$\pm$0.07 & 93 \\
      group & 0.57$\pm$0.04 &  0.33 & 0.68$\pm$0.04 & 0.65$\pm$0.03 & 180 \\
      product & 0.45$\pm$0.12 &  0.20 & 0.53$\pm$0.07 & 0.44$\pm$0.16 & 144 \\
      creative-work & 0.63$\pm$0.03 & 0.16 & 0.75$\pm$0.02 & 0.55$\pm$0.02 & 184  \\ \hline
    \end{tabular}
    \caption{16-shot NER F1-scores and standard deviations on the WNUT2017 dataset. The third column shows the results obtained with the baseline UH-RiTUAL.}
    \label{tab:ner2}
\end{small}
\end{table*}

\begin{table}
\begin{small}
    \centering
    \begin{tabular}{c|c|cc} 
     \textbf{Dataset} & \textbf{NE Type} & \textbf{F1} & \textbf{Count} \\\hline
     CoNLL-2003 & person & 0.74$\pm$0.09 & 1537  \\
     & location & 0.79$\pm$0.01 & 1899 \\
    & organisation  & 0.73$\pm$0.01 & 1843\\ \hline
    MIT Movie &  person  & 0.80$\pm$0.04 & 1908 \\
   & creative work & 0.43$\pm$0.08 & 906 \\ \hline
    \end{tabular}
    \caption{16-shot NER F1-scores and the standard deviations on CoNLL-2003 and MIT Movie datasets.}
    \label{tab:ner}
\end{small}
\end{table}

\subsection{NET in a Zero-shot Setup} 
Tables \ref{tab:entitytyping} and \ref{tab:dbp_entitytyping} show that GPT2 without relying on any NE context or other resources can perform NET quite well on most NE lists.
On CoNLL-2003, the results are even close to those obtained by ZOE \cite{zhou-etal-2018-zero}, a much more complex system.
Higher scores are obtained for regular types such as \textit{person} or clean NEs (e.g. DBpedia NEs).
We see lower scores for \textit{creative work} in MIT Movies and \textit{location} in CoNLL-2003, these being often confused with \textit{person} (in 53\% of the cases) and \textit{organisation} (in 29\% of the cases) respectively.
The first confusion is not surprising given that movie titles could contain character names while \textit{character} is included in \textit{person}. 
The second confusion, \textit{location}-\textit{organisation}, is already mentioned as a common issue \cite{derczynski-etal-2017-results}. 
Thus, most likely including context would help to disambiguate such NEs that belong to multiple types.

The NET performance on memorized NEs is, as expected, larger than on unmemorized NEs.
However, Table \ref{tab:entitytyping} shows a much smaller drop on MIT Movie than on CoNLL-2003. 
The difference between these NE lists lies in the criterion we used in the memorization test, either focused on knowing NE's individual words in CoNLL-2003 or transitions between words in MIT Movie. 
This suggests that for a better NET performance, \textit{the model's exposure to the NE's words weighs much more than the exposure to the exact NE, i.e. its word transitions}. 
In other words, even if a specific NE was not seen during pre-training, but its composing words were present as part of other NEs, then the model could still leverage this exposure in order to correctly classify  the unseen NE in the proposed setup.  
This is further confirmed on the larger DBpedia NE lists (Table \ref{tab:dbp_entitytyping}).

Table \ref{tab:wnut_entitytyping} shows that the model in a zero-shot setup yields significantly higher results than the random baseline on WNUT2017 NEs;
though, overall lower for this textual genre,
except for \textit{person} and \textit{location}. 
The Twitter-style NEs may contain many words unseen by the model during pre-training.
Also, we noticed similar confusion patterns as before: \textit{corporation} or \textit{group} (associated with \textit{organisation}) with \textit{location}, and \textit{creative work} with \textit{person}.
Thus, context seems again promising with both the typing of NEs with new / rare words and the disambiguation for related NE types.

\subsection{NER in a Few-shot Setup} 
As presented in Tables \ref{tab:ner2} and \ref{tab:ner}, the pre-trained LM, without any further fine-tuning or training can perform NER surprisingly well in the designed few-shot settings, even on noisy data.
On WNUT2017, the noisiest dataset, we can see that the model outperforms the supervised baseline for all NE types apart from \textit{location}.
Similar to the baseline, \textit{location} and \textit{person} are among the easiest to extract NE types, while \textit{product} is quite hard.
In contrast, \textit{corporation} and \textit{creative work} types are recognized rather well.
A qualitative analysis of the predicted NEs for CoNLL-2003 shows that the model has more challenges with false positives.
This suggests that more negative examples may be needed at inference.
For MIT Movie, the model often predicts  "none" for \textit{creative work}, an issue that might be overcome with better chosen positive examples. 

\textbf{Test (un)seen} in Table \ref{tab:ner2} shows F1-scores when all NEs are replaced by random strings (lists available in Appendix), while fixing the context and the query examples. 
The scores for \textbf{Test as-is} are lower than for  \textbf{Test seen} and larger than for \textbf{Test unseen}.
Also, the score differences between \textbf{Test seen} and \textbf{Test as-is} are larger than the ones between \textbf{Test unseen} and \textbf{Test as-is}.
These results lead to the rejection of hypothesis 1 and confirmation of hypothesis 2 introduced at the end of Section \ref{sec:few} and show that \textit{the model appears to prioritize NEs cues more than context cues in few-shot settings}.
Thus, when choosing query examples, one may favour to focus more on providing diverse NE patterns for an entity type than diverse context patterns.

As for NET, the impact of \textit{NE (un)memorization during pre-training} is significant, with some exceptions such as \textit{product}, for which F1-scores on \textbf{Test as-is} are almost the same to F1-scores on \textbf{Test unseen}.
We checked the NEs sampled for the query and noticed frequent irregular names for \textit{product}.
Previously, \newcite{lin-etal-2020-rigorous} showed that name regularity is critical for generalization to new NEs.
However, while the placeholders we used for unmemorized NEs were highly irregular, \textit{this irregularity when often present for a certain NE type appears an exploitable cue}.

The best F1-scores reported by \newcite{huang2020few} are in the range $0.65$-$0.90$ on the original CoNLL-2003, $0.56$-$0.67$ on the MIT Movie with the original NE types and $0.38$-$0.51$ on the original WNUT2017, depending if 5 examples or 10\% of the data are used as shots.
Our average F1-scores with 16 shots appear competitive on our sub-sampled test sets: $0.75$ for CoNLL-2003, $0.62$ for MIT Movie and 0.61 for WNUT2017, which appears to validate the effectiveness of the pre-trained GPT2 probed for NER under the proposed conditions.

\section{Conclusion}
\label{sec:conclusion}
We proposed a NET and NER probing methodology designed for pre-trained auto-regressive LMs in zero- or few-shot settings.
Our goal was to investigate if such a model, without any fine-tuning, could handle the tasks well while generalising to noisy text, diverse NE types, and new NEs.
For this, we also defined a novel procedure to assess the exposure of the model to various NEs to create sets of (un)memorized NEs.
Overall, we deemed our setup under more realistic conditions inspired by human linguistic behavior.

The results showed that a medium-size GPT2 in the proposed settings was quite good at NET and NER and we revealed multiple new insights, impactful for future work.

With pre-trained encoders, the exposure of the LM to NEs should not be investigated in fine-tuning only while neglecting the memorization during pre-training.
We have proposed an effective method to support future studies with this.
Also, a LM already pre-trained on a general task and a large corpus could effectively bootstrap NER for new applications, especially when NEs are common constructs in a language.
Frequent name irregularity for a type in context can become a regularity effectively exploited by the LM in a few-shot NER. 
Context is important but with limited impact in our studied setup.
Finally, choosing good query examples of NE patterns in context for few-shot NER and extending the study to other auto-regressive or masked LMs are still matters of investigation.

 \appendix

\section*{Appendix: Additional experiment details}
\label{app:details}

In experiments, we used an NVIDIA GTX 1080 with 11GB RAM.
We show the running time for each experiment in Table \ref{tab:runtime}.
\begin{table}[hbt!]
    \begin{tabular}{l|c|cc}
      \textbf{Dataset} & \textbf{NE type} & \multicolumn{2}{c}{\textbf{Time(s)}}\\
      & & \textbf{NET} & \textbf{NER} \\ \hline
      CoNLL-2003 & person & 210 & 6371 \\
       & location & 85 & 6906 \\
       & organisation & 144 & 6202\\
       \hline
      MIT Movie & person & 131 & 4850\\
      & creative work & 100 & 2713\\\hline
      WNUT2017 & person & 86 & 2759 \\
      & location & 49 & 1081\\
      & corporation & 29 & 730\\
      & group & 40 & 1372\\
      & product & 38 & 1127 \\
      & creative work & 39 & 1502\\ \hline
    \end{tabular}
    \caption{Running time in seconds for each experiment.}
    \label{tab:runtime}
\end{table}

Lists of memorized and unmemorized NEs used to create the \textit{test seen} and \textit{test unseen} datasets for assessing NER generalization are presented further.

Memorized:
\begin{itemize}
    \item \textit{person}: Mary, Steve, Davis, Sam, Robert, Alex, Michelle, James, Danny, Rose, Edward, Rob, Harry, Tom, Paul
    \item \textit{location}: Florida, Toronto, Syria, India, Houston, America, France, Australia, Turkey, NEW YORK, Chicago, Germany, Scotland, Washington, Ukraine
    \item \textit{corporation}: Reuters, CNN, NBA, Uber, YouTube, CBC, Netflix, Microsoft, Twitter, Facebook, Apple, MAC, Tesla, Disney, Reddit
    \item \textit{group}: Army, Chicago Blackhawks, Real Madrid, CIA, Senate, ART, NBA, The Black Keys, Crystal Palace, European Union, green day, Labor, Chelsea, the warriors, Democrats
    \item \textit{product}: Air Music Jump, Android, Linux OS, iOS, Windows 7, Tesla, Google Music, SQL, Amazon Prime, Nintendo plus, google pixel, iPhone, Xbox 360, Legendary Skin, Bio Spot
    \item \textit{creative work}: Black Swan, Iron Man 2, Finding Bigfoot, Good Morning Britain, Teen Titans, Pac- Man, Game of Thrones, La La Land, Last Christmas, Star Wars, Doctor Who, the Twilight Zone, Pokémon, Star Trek, Minecraft
\end{itemize}
Unmemorized:
\begin{itemize}
    \item all: xgwqicng, kiooaiql, wpvqymid, rrmihdcg, owblmgbx, tiybjelq, ytlbllnh, ybwifxxv, svlsskxx, jdtqyoov, tzrtffbu, jvwywjhy, hzhwhahw, gjrmquke, gmenqwpb
\end{itemize}

The thresholds set for pruning the exposure metrics in order to select \textit{memorized NEs} and \textit{unmemorized NEs} are presented in Tables \ref{tab:thresholds1} and \ref{tab:thresholds2}.

\begin{table}[hbt!]
    \centering
    \begin{tabular}{c|cc}
      \textbf{Dataset} & $\text{exposure}_{\theta}^{word}$ & $\text{exposure}_{\theta}^{trans}$ \\ \hline
      DBpedia & 1 & - \\
       & - & 0.01 \\ \hline
     % & W, T & 0.2 & 0.05 \\\hline
     CoNLL & 0.8 & - \\ \hline
     MIT Movie & - & 0.001  \\ \hline
    \end{tabular}
    \caption{Thresholds used for pruning the exposure metrics in order to select \textit{memorized NEs}.}
    \label{tab:thresholds1}
\end{table}

\begin{table}[hbt!]
    \centering
    \begin{tabular}{c|cc}
      \textbf{Dataset} & $\text{exposure}_{\theta}^{word}$ & $\text{exposure}_{\theta}^{trans}$ \\ \hline
      DBpedia & 1e-06 & - \\ \hline
       & - & 1e-06 \\ 
     %   & 1e-06 & 1e-06 \\\hline
     CoNLL & 1e-04 & -\\ \hline
     MIT Movie & - & 1e-05  \\ \hline
    \end{tabular}
    \caption{Thresholds used for pruning the exposure metrics in order to select \textit{unmemorized NEs}.}
    \label{tab:thresholds2}
\end{table}

% \nocite{*}
\section{Bibliographical References}\label{reference}
%\label{main:ref}

\bibliographystyle{lrec2022-bib}
\bibliography{anthology,custom}

\end{document}